%% file: main.tex
\documentclass[11pt]{article}
\usepackage[]{acl}
\usepackage{latexsym}
\usepackage[T1]{fontenc}
\usepackage{lmodern}
\usepackage[utf8]{inputenc}
\usepackage{booktabs}
\usepackage{graphicx}
\usepackage{amsmath}
\usepackage{amssymb}
\PassOptionsToPackage{hyphens}{url}
\usepackage{url}
\title{Does Chain-of-Thought Reliance Follow a Scaling Law?\\
       Larger Models Follow Their Steps More, Independent of Post-Training}
\author{
  Esther Xin \\
  Independent Researcher \\
  \texttt{estherxin0011@gmail.com}\\
  {\small Code and data: \url{https://github.com/ethxin0011/cot-scale-reliance}}
}

\begin{document}
\maketitle

\begin{abstract}
Monitoring a model's chain of thought (CoT) is useful for safety only if the
final answer actually depends on the written steps. Prior measurements of this
dependence use natural tasks on which the model's true reasoning is unknown. We
measure it against an exact ground truth. On a synthetic box-swap task every
intermediate state is known, so we can replace one step inside a model's own
chain with a wrong but valid state and check whether the continuation and the
answer follow the edit. We evaluate 25 open-weight models---Qwen3 (0.6B--14B,
base and post-trained), Gemma~3 (1B--12B), Llama~3 (1B--8B) and eight OLMo~2
pretraining checkpoints---on 7{,}200 items each, using a single 16\,GB GPU. We
find five results.
\textbf{(1) Reliance rises with size and saturates.} In Qwen3 the answer follows
an edited step 16--18\% of the time at 0.6B and 97--100\% at 14B; a saturating
logistic in $\log N$ fits with midpoint near 1B ($R^2=0.89$), and Gemma~3 and
Llama~3 rise the same way.
\textbf{(2) Capability is the main driver.} Across all 25 models, reliance and
step accuracy correlate at $r=0.99$; equally capable models are about equally
reliant, whatever their size or family.
\textbf{(3) Size still matters at matched correctness.} Counting only problems the
model solved correctly before the edit, reliance still increases with size among
capable models ($\rho=0.80$, $p=0.002$).
\textbf{(4) Post-training does not change reliance.} Base and post-trained Qwen3
differ by at most 0.03 from 4B upward; the one large gap (1.7B) coincides with a
large capability gap.
\textbf{(5) No model up to 14B solves the task internally.} Without written steps,
every capable model is at chance once a problem has four or more swaps.
OLMo~2 checkpoints never become capable on the task, so the training-token axis
is reported as a capability result and a limitation. On tasks that need
step-by-step computation, scale makes the chain of thought more load-bearing,
not less, which supports monitoring it in that regime.
\end{abstract}

\section{Introduction}
\label{sec:intro}

Chain-of-thought prompting \citep{wei2022cot} turns hidden computation into
visible text, and reading that text has become a candidate safety tool
\citep{korbak2025monitorability, baker2025monitoring}. The tool only works if the
written steps are \emph{load-bearing}: if the model's answer would be the same
whatever the steps said, the chain is a narrative rather than a record. Measuring
this property, often called CoT faithfulness, is hard on natural tasks because
the model's true reasoning is not observable. Existing evidence is mixed.
\citet{lanham2023faithfulness} intervene on chains (truncation, inserted
mistakes, paraphrase) and find that, on many tasks, larger models depend
\emph{less} on their chains; \citet{bentham2024disguised} show that this
measure is strongly correlated with accuracy and that much of the apparent
unfaithfulness of small models disappears once answer bias is controlled.
\citet{turpin2023unfaithful} and
\citet{chen2025reasoning} show that stated reasoning can omit factors that
demonstrably drive the answer. \citet{emmons2025necessary} argue that the picture
changes when the chain is \emph{necessary} for the task.

These results leave open a basic scaling question: as models grow, do they rely on
their written steps more or less, and is this governed by size, training data, or
post-training? Answering it needs three things existing work lacks at once:
(i) a ground truth for every intermediate step, so an edit's effect is
unambiguous; (ii) a task on which the chain is genuinely needed, so reliance is
not trivially zero; and (iii) many models that vary one factor at a time.

We provide all three with a deliberately simple design. Models track four
lettered boxes through $k$ position swaps, writing the full state after each swap.
Because every state is determined, we can (a) score each written step exactly,
(b) measure accuracy with and without the chain, and (c) replace one step inside
the model's own chain with a wrong but valid state and let it continue. The
\emph{follows-edit rate}---how often the continuation and the answer are
consistent with the corrupted state---is an exact, item-level measure of how
much the answer depends on what was written.

\paragraph{Findings.}
\textbf{(1) Reliance scales with size and saturates.} In all three families with
size series, the follows-edit rate rises monotonically with parameters; in Qwen3
it fits a saturating logistic in $\log N$ with midpoint near 1B and is at
0.97--1.00 by 14B (Figure~\ref{fig:size}, Table~\ref{tab:fits}).
\textbf{(2) Capability is the operative variable.} Across all 25 models, reliance
and step accuracy are almost perfectly correlated ($r=0.99$); at matched
capability, a 1.7B Qwen3, a 4B Gemma~3 and an 8B Llama~3 are about equally
reliant (Figure~\ref{fig:capability}). This confirms, with an exact ground truth,
the accuracy--faithfulness coupling reported by \citet{bentham2024disguised}.
\textbf{(3) Size matters beyond correctness.} Restricting to problems the model
solved correctly before the edit, reliance still increases with size among
capable models ($\rho=0.80$, $p=0.002$).
\textbf{(4) Post-training is not the driver.} Base and post-trained Qwen3
checkpoints are within $\pm0.03$ in reliance from 4B upward; the one large gap
(1.7B) coincides with a large capability gap (Table~\ref{tab:post}).
\textbf{(5) No internal shortcut up to 14B.} Without a chain, every capable model
is at chance (mean 0.041 vs.\ $1/24=0.042$) for $k\ge4$; only two-swap problems
show size-dependent internal solving (0.09 to 0.58 in Qwen3).
The OLMo~2 pretraining checkpoints (1B and 7B, 0.3T--4T tokens) never reach
the capability threshold, so the training-token axis is reported as a
capability result and a limitation.

\paragraph{Contributions.} (i) An exact, ground-truth protocol for measuring CoT
reliance by in-chain step edits, runnable on a single 16\,GB GPU; (ii) to our
knowledge, the first scaling study of CoT reliance measured against an exact
ground truth, with matched base and post-trained checkpoints at every size;
(iii) evidence that, on tasks requiring
serial computation, reliance follows capability and saturates rather than
eroding with scale, reconciling this regime with prior inverse-scaling
observations; and (iv) released items, generations and analysis code.

\section{Related Work}
\label{sec:related}

\paragraph{Measuring CoT faithfulness.}
\citet{lanham2023faithfulness} test whether answers change when chains are
truncated, corrupted or paraphrased, and report that larger models are often
less faithful, an inverse-scaling pattern. \citet{bentham2024disguised}
replicate this setup across three model families and find that, after
normalising for a model's bias toward particular answer choices, small models
look far less unfaithful and the normalised metric is strongly correlated with
accuracy ($R^2=0.74$), raising doubts about whether it measures faithfulness or
capability. Our capability analysis (Section~\ref{sec:capability}) addresses
this concern directly, with an exact ground truth and a correct-prefix control.
\citet{turpin2023unfaithful} show that
biasing features can steer answers without being mentioned, and
\citet{chen2025reasoning} extend this to reasoning models; across twelve
open-weight reasoning models, \citet{young2026lietome} find that training
method and model family predict hint acknowledgment more strongly than parameter
count, and \citet{hase2026cst} find that larger models are not more faithful
out of the box. These works measure whether a chain \emph{discloses} an
influencing cue, a different property from whether the answer \emph{depends}
on the written steps. \citet{paul2024frodo}
measure and train for causal use of intermediate steps, \citet{tutek2025fur}
measure parametric faithfulness by unlearning reasoning steps, and
\citet{ye2026nldd} corrupt individual steps and measure the drop in answer
confidence, finding that late steps matter little. These works use natural
or benchmark tasks where the correct intermediate reasoning is not uniquely
defined, so an edit's ``correct'' downstream effect must be inferred. Our task
removes that ambiguity: every intermediate state and every consequence of an
edit is computable.

\paragraph{Robustness to corrupted steps.}
\citet{aravindan2026fragile} inject errors into mathematical chains for 13 models
and find that larger models lose much less accuracy, i.e.\ they are more robust
to corrupted steps. This is a different quantity from ours and, as we discuss in
Section~\ref{sec:discussion}, the two results are compatible.

\paragraph{Monitorability and necessity.}
\citet{korbak2025monitorability} frame CoT monitoring as a valuable but fragile
safety opportunity, and \citet{baker2025monitoring} show that optimisation
pressure on chains can produce obfuscation. \citet{wang2026monitorbench} find in
MonitorBench that more capable models tend to be less monitorable, but that
monitorability is higher when the decision-critical factors shape the
intermediate reasoning rather than only the final answer.
\citet{emmons2025necessary} argue that
when a chain is necessary for a task, models struggle to evade monitors, and
\citet{pfau2024dots} show that some computation can be moved into filler tokens.
Our results give a quantitative, scaling-level version of the necessity argument:
on a task where the direct answer is at chance, reliance increases with scale.

\paragraph{Why CoT helps on serial tasks.}
Theory shows that chains extend what fixed-depth transformers can compute,
especially for inherently serial problems
\citep{feng2023mystery, merrill2024expressive, li2024serial}. Permutation and
entity tracking is a canonical example; \citet{kim2023entity} study box-content
tracking in language models. We use this class of task precisely because direct
computation should fail as $k$ grows, which we confirm empirically.

\paragraph{Scaling.}
Scaling laws describe loss as a function of parameters and data
\citep{kaplan2020scaling, hoffmann2022chinchilla}; downstream abilities can appear
abruptly \citep{wei2022emergent}, though apparent emergence can depend on the
metric \citep{schaeffer2023mirage}. We report both a continuous reliance measure
and its relation to capability, so that a jump in one can be checked against the
other.

\section{Method}
\label{sec:method}

\subsection{Task}
\label{sec:task}
A problem consists of an initial arrangement of letters A--D in boxes 1--4 and a
list of $k$ swaps of box positions, $k\in\{2,4,8,12,16,24\}$. We generate 300
problems per $k$ (1{,}800 in total). The prompt contains an instruction and three
worked examples ($k=2,3,4$) in the exact target format, in which every step
restates its swap and the full state:
\texttt{Step 3 (swap 1 and 2): 1:A 2:B 3:C 4:D}, followed by a final
\texttt{Answer:} line. With four letters there are $4!=24$ final states, so
chance accuracy is $1/24\approx0.042$. Earlier pilots with five unlabelled boxes
produced near-zero step accuracy for all small models; labelled positions and
restated swaps were adopted so that the task measures state tracking rather than
format following.

\subsection{Conditions}
\label{sec:conditions}
Each problem is rendered in four conditions; three are analysed here.
\textbf{CoT}: the model writes all steps and the answer.
\textbf{Direct}: the model must output the answer line only.
\textbf{Edit}: the prompt includes the start of a solution, steps $1..m$
($1\le m<k$), in which the state at step $m$ is replaced by a different valid
arrangement; the model continues from step $m+1$. A fourth control condition is
generated for every problem but not analysed in this paper.

\subsection{Metrics}
\label{sec:metrics}
A line-based parser reads each output up to the first answer line and discards
malformed states (for example, a letter appearing twice).
\emph{Compliance} is the fraction of outputs with well-formed steps and an answer.
\emph{Prefix accuracy}, our capability measure, is the fraction of CoT runs whose
states are correct up to the edit position. \emph{CoT} and \emph{direct accuracy}
are final-answer accuracies. Each edited run is classified as
\emph{follows-edit} (the continuation and answer equal the result of applying the
remaining swaps to the corrupted state), \emph{repair} (it returns to the true
trajectory), \emph{wrong-other}, or \emph{invalid}. The \emph{follows-edit rate}
is our reliance measure. Because a model that cannot track the state may fail to
follow an edit for reasons unrelated to reliance, we also report it
\emph{conditional on a correct prefix}, i.e.\ on problems where the model's own
unedited chain was correct up to step $m$.

\subsection{Models}
\label{sec:models}
\textbf{Qwen3} \citep{qwen3}: 0.6B, 1.7B, 4B, 8B and 14B, each as a base and a
post-trained checkpoint (10 models), trained on 36T tokens at every size; the
post-trained models are run with thinking disabled.
\textbf{Gemma~3} \citep{gemma3}: 1B, 4B and 12B instruction-tuned.
\textbf{Llama~3} \citep{llama3}: 3.2-1B, 3.2-3B and 3.1-8B instruction-tuned.
\textbf{OLMo~2} \citep{olmo2}: 1B checkpoints at 336B, 944B, 1972B and 2978B
tokens plus the final release, and 7B checkpoints at 1934B and 2924B tokens plus
the final release. OLMo~2 is the only family that releases intermediate
checkpoints of the same model, which is what a training-token axis requires: in
Gemma~3 and Llama~3 token count rises together with size, and in Qwen3 it does
not vary. Size $N$ is the non-embedding parameter count. A 4-bit NF4
\citep{dettmers2023qlora} copy of Qwen3-4B serves as a quantization ablation.

\subsection{Inference and statistics}
\label{sec:inference}
All runs are inference-only on one NVIDIA T4 (16\,GB). Models that do not fit in
16\,GB at 16-bit precision are loaded with NF4 weights. Generation is greedy and
stops at the first answer line, which does not affect scoring because the parser
ignores text after it. A checkpointed runner survives preemption, and every model
in the paper has all 7{,}200 generations. Confidence intervals use 1{,}000 bootstrap
resamples over problems. We fit
$y=\ell+(1-\ell)\,\sigma\!\left(a(\log_{10}N-b)\right)$ to each metric and report
Spearman correlations with $N$. A model counts as \emph{capable} if its prefix
accuracy is at least 0.5, a threshold fixed before the large-model results were
inspected. Full settings are in Table~\ref{tab:config}.

\section{Results}
\label{sec:results}

\input{tables}

\subsection{The task requires the chain}
\label{sec:need}
\begin{figure*}[!htbp]
\centering
\includegraphics[width=0.95\textwidth]{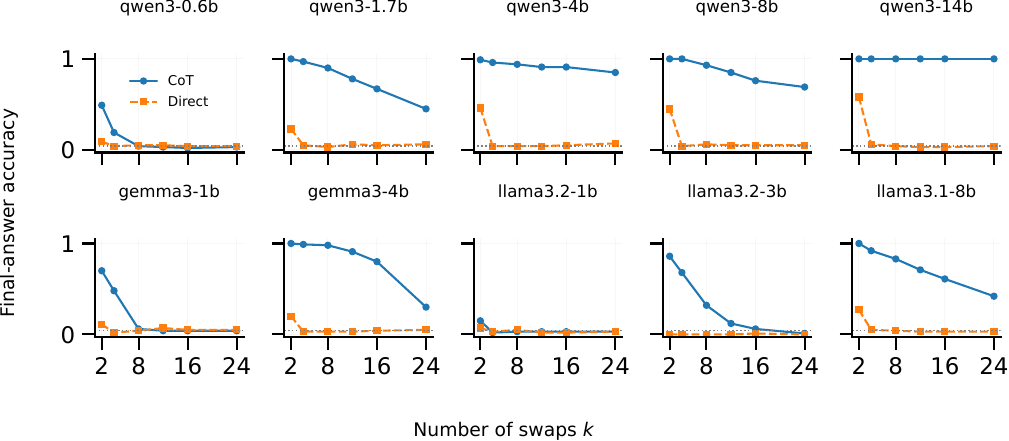}
\caption{Final-answer accuracy with (CoT) and without (Direct) the written chain,
by number of swaps $k$, for the post-trained models. The dotted line is chance
($1/24$). Direct accuracy is at chance for $k\ge4$ at every size.}
\label{fig:byk}
\end{figure*}

Figure~\ref{fig:byk} shows final-answer accuracy by $k$. With a chain, capable
models stay far above chance up to $k=24$ (Qwen3-14B: 1.00 at every $k$). Without
one, every model is near chance from $k=4$: across the 12 capable models in the
three families, direct accuracy for $k\ge4$ ranges from 0.002 to 0.052 with mean
0.041 (chance 0.042), and for the largest models it is not significantly above
chance (binomial $p=0.61$ for Qwen3-14B, $p=0.11$ for Gemma~3-12B). Only at $k=2$
does direct accuracy grow with size (Qwen3 post-trained 0.09 to 0.58; Spearman
$\rho=0.82$ with $N$ over the three families). The models therefore start to
compute two swaps internally, but none of them, up to 14B, can do four without
writing the steps.

\subsection{Reliance rises with size and saturates}
\label{sec:size}
\begin{figure}[!htbp]
\centering
\includegraphics[width=0.62\textwidth]{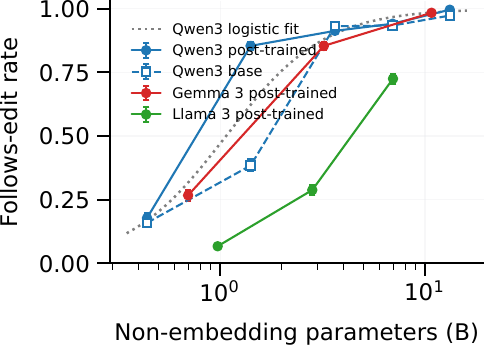}
\caption{Follows-edit rate vs.\ non-embedding parameters, with 95\% bootstrap
CIs. Filled markers: post-trained; open markers: base. Dotted curve: saturating
logistic fit to the ten Qwen3 models.}
\label{fig:size}
\end{figure}

Table~\ref{tab:main} and Figure~\ref{fig:size} report the follows-edit rate. In
Qwen3 it rises from 0.16 (base) and 0.18 (post-trained) at 0.6B to 0.97 and 1.00
at 14B. A saturating logistic in $\log N$ fits the ten Qwen3 models with
$R^2=0.89$ and a midpoint of 1.07B; the fitted floor is 0.00
(Table~\ref{tab:fits}). The same monotone increase appears in Gemma~3
(0.27, 0.85, 0.98 at 1B, 4B, 12B) and Llama~3 (0.07, 0.29, 0.73 at 1B, 3B, 8B).
Pooling the three families, Spearman $\rho=0.90$; the pooled logistic fit is
weaker ($R^2=0.74$) because families reach the same reliance at different sizes.
Edited models rarely repair the corruption: repair is at most 0.06 for any model
(Table~\ref{tab:edits}), and answers that contradict the model's own final step
are at most 0.02. When capable models do not follow an edit, they mostly produce
a different wrong trajectory, not the correct answer.

\subsection{Capability is the operative variable}
\label{sec:capability}
\begin{figure}[!htbp]
\centering
\includegraphics[width=0.62\textwidth]{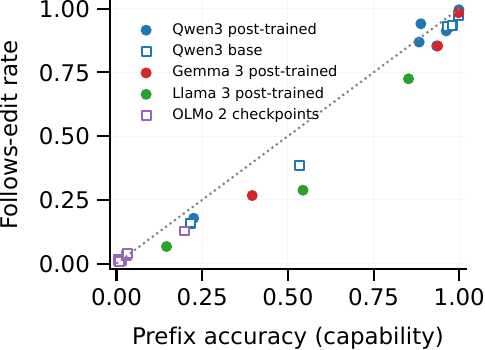}
\caption{Follows-edit rate vs.\ prefix accuracy for all 25 models. Reliance and
capability are almost perfectly correlated across families and training stages.}
\label{fig:capability}
\end{figure}

Figure~\ref{fig:capability} plots reliance against prefix accuracy for all 25
models. The points collapse onto one curve (Pearson $r=0.99$, Spearman
$\rho=0.98$), including the OLMo~2 checkpoints, which sit at the low end. Models
of different sizes and families with similar capability have similar reliance:
Qwen3-1.7B (prefix 0.93, reliance 0.85), Gemma~3-4B (0.94, 0.85) and
Llama~3.1-8B (0.85, 0.73). Size matters for reliance mainly through capability;
families differ in how many parameters they need to reach a given capability.
This confirms and sharpens the observation of \citet{bentham2024disguised} that
a dependence-based faithfulness measure tracks accuracy ($R^2=0.74$ in their
setting): with an exact ground truth for every step, the coupling is
near-perfect.

A tight coupling could arise mechanically, since a model that cannot track the
state cannot propagate an edit either. The conditional measure addresses this.
Restricting to problems on which the model's own prefix was correct, reliance is
higher for small models (Qwen3-0.6B: 0.44 vs.\ 0.18 unconditional) but still
increases with size: among the 12 capable models, Spearman $\rho=0.80$
($p=0.002$), and in Qwen3 the conditional curve fits a logistic with $R^2=0.91$.
Even when a small model gets the steps right, its answer depends on them less
than a large model's does. Reliance is therefore not merely accuracy in
disguise: the size trend survives once correctness is held fixed.

\subsection{Post-training does not change reliance}
\label{sec:post}
\begin{figure}[!htbp]
\centering
\includegraphics[width=0.62\textwidth]{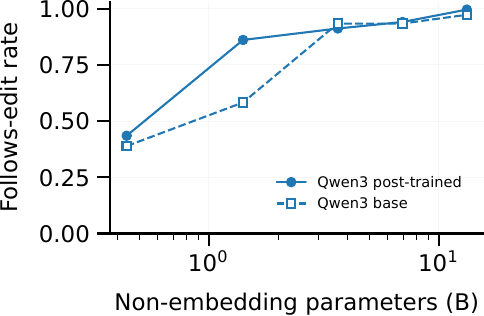}
\caption{Qwen3 base vs.\ post-trained follows-edit rate, restricted to problems
on which the model's own unedited prefix was correct.}
\label{fig:post}
\end{figure}

Table~\ref{tab:post} and Figure~\ref{fig:post} compare base and post-trained
Qwen3 at each size, conditional on a correct prefix. From 4B upward the two are
within 0.03 ($-0.022$, $+0.008$, $+0.022$). The one large difference, $+0.278$
at 1.7B, coincides with a large capability difference (prefix accuracy 0.533 vs.\
0.934), consistent with Section~\ref{sec:capability}. On this task, post-training
raises reliance only by raising capability where the base model is still weak.

\subsection{Quantization}
\label{sec:quant}
Loading Qwen3-4B in NF4 lowers prefix accuracy from 0.962 to 0.883 and reliance
from 0.913 to 0.869 (Table~\ref{tab:main}). Because the largest models are
quantized, any bias from quantization works against the size trend, so the
reported increase is, if anything, conservative.

\subsection{Training tokens: a capability floor}
\label{sec:tokens}
\begin{figure}[!htbp]
\centering
\includegraphics[width=0.62\textwidth]{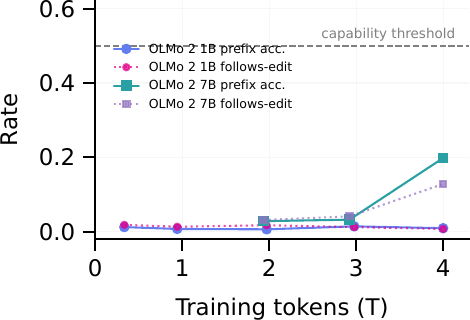}
\caption{OLMo~2 prefix accuracy and follows-edit rate across pretraining
checkpoints. The final release includes mid-training. No checkpoint crosses the
0.5 capability threshold (dashed).}
\label{fig:olmo}
\end{figure}

Table~\ref{tab:olmo} and Figure~\ref{fig:olmo} report the OLMo~2 checkpoints. All
five 1B checkpoints are at the floor (prefix accuracy $\le0.014$) across a 12-fold
increase in training tokens. At 7B, the 1.9T and 2.9T checkpoints are also at the
floor (0.028, 0.032), and only the released checkpoint rises, to 0.198, with
direct accuracy at $k=2$ rising to 0.217. That checkpoint differs from the 2.9T
one not only in tokens but also in OLMo~2's math-heavy mid-training stage
\citep{olmo2}, so we cannot attribute the jump to token count. No checkpoint
reaches the capability threshold, so the token axis cannot support a reliance
scaling curve. The one checkpoint with enough conditional data
(\texttt{olmo2-7b-main}, $n=356$) has conditional reliance 0.303, within the range
of similarly capable models from other families (0.14--0.49 for prefix accuracy
0.15--0.40), consistent with the capability account.

\clearpage
\section{Discussion}
\label{sec:discussion}

\paragraph{Reconciling with inverse scaling.}
\citet{lanham2023faithfulness} found that larger models often depend less on their
chains, and \citet{wang2026monitorbench} find that more capable models tend to be
less monitorable. Our results do not contradict this; they locate a regime where the
opposite holds. When a task is within a model's internal capacity, the chain can
be redundant and larger models may answer correctly whatever it says. When the
task needs more serial steps than the model can do internally---here, every
$k\ge4$ problem for every model up to 14B---the chain is the computation, and
larger models use it more consistently. This matches MonitorBench's own
observation that monitorability is higher when decision-critical factors shape
the intermediate reasoning. The $k=2$ result marks the boundary: as
direct accuracy rises with size there, we would expect reliance on short chains
to eventually fall. The scaling of reliance should therefore be stated relative
to the task's difficulty for the model, which supports the necessity argument of
\citet{emmons2025necessary}.

\paragraph{Reconciling with robustness to corrupted steps.}
\citet{aravindan2026fragile} report that larger models are \emph{more} robust to
errors injected into mathematical chains, which might seem to contradict our
finding that larger models follow edits more often. The two settings differ in
what a correct continuation looks like. An arithmetic error in a math chain
usually contradicts information that is still present in the problem, so a
capable model can recompute the affected quantity and recover; robustness there
reflects re-derivation from the question. Our edit replaces the entire state with
a different \emph{valid} arrangement, so the only way to ``recover'' is to
silently re-derive every preceding swap from the start while ignoring the written
state. Following the edit is the behaviour that reflects genuine reliance on the
chain, and it is what larger models do. Robustness to local errors and reliance on
the written state are therefore distinct properties, and both can rise with scale.

\paragraph{Implications for monitoring.}
For tasks that genuinely need multi-step computation, the evidence here favours
CoT monitoring: up to 14B, the more capable the model, the more its answer is
determined by what it wrote, and edits are almost never silently repaired. The
caveat is the open end of the curve. Our largest models saturate on our metric,
so the data cannot show whether reliance would eventually fall once models can do
long chains internally. The direct-accuracy curve at small $k$ is the variable to
watch.

\paragraph{What a ``scaling law'' means here.}
Reliance is well described by a saturating function of $\log N$ within a family,
but the more fundamental regularity is its relation to capability, which holds
across families, sizes and training stages. A useful predictive form is therefore
two-stage: scale predicts capability on the task, and capability predicts
reliance.

\section{Limitations}
\label{sec:limitations}
\textbf{One task family.} Box-swap tracking is a clean, controllable serial task;
reliance on natural-language, mathematical or agentic reasoning may scale
differently. \textbf{Size range and saturation.} Models go up to 14B, and the
largest saturate near 1.0, which limits what the top of the curve can show.
\textbf{Token axis.} OLMo~2 checkpoints up to 7B never reach the capability
threshold, so we cannot estimate how reliance scales with training tokens; the
13B OLMo~2 checkpoints were not run within our compute budget. \textbf{Family
confounds.} Across families, data, architecture and recipe differ together; we
therefore draw size conclusions within families and use pooling only as
corroboration. \textbf{Behavioural measure.} Following an edit shows that the
answer depends on the written state; it does not show that no parallel internal
computation occurs. \textbf{Single edit type and greedy decoding.} We edit one
step to one wrong valid state and decode greedily; other edit types or sampling
may give different rates. \textbf{Quantization.} Large models run in NF4; the
4B ablation suggests a small downward effect. \textbf{Reasoning modes.} Qwen3's
thinking mode was disabled; models trained to reason in long hidden traces were
not studied.

\section{Conclusion}
\label{sec:conclusion}
We measured how much a model's answer depends on its own chain of thought, using
exact in-chain step edits on a task with known intermediate states, across 25
open models from four families on a single 16\,GB GPU. The results support five
conclusions.

\begin{enumerate}
\item \textbf{Reliance on the chain rises with model size and saturates.} The
follows-edit rate climbs from 0.16--0.18 at Qwen3-0.6B to 0.97--1.00 at
Qwen3-14B, fits a saturating logistic in $\log N$ with midpoint near 1B
($R^2=0.89$), and rises monotonically in Gemma~3 (0.27, 0.85, 0.98) and Llama~3
(0.07, 0.29, 0.73) as well.
\item \textbf{Capability, not size itself, is the main driver.} Across all 25
models, reliance and step accuracy correlate at $r=0.99$. Families need different
numbers of parameters to reach a given capability, but at matched capability
their reliance is similar.
\item \textbf{Size still matters once correctness is controlled.} On problems the
model solved correctly before the edit, reliance still increases with size among
capable models ($\rho=0.80$, $p=0.002$), so the trend is not only an artefact of
small models failing the task.
\item \textbf{Post-training does not change reliance.} Base and post-trained Qwen3
differ by at most 0.03 from 4B upward; the only large gap, at 1.7B, matches a large
capability gap (prefix accuracy 0.53 vs.\ 0.93).
\item \textbf{No model up to 14B solves the task without its chain.} Direct
accuracy is at chance (0.041 vs.\ 0.042) for every capable model once a problem
has four or more swaps; internal solving appears only for two-swap problems
(0.09 to 0.58 in Qwen3).
\end{enumerate}

\noindent The OLMo~2 checkpoints never reached the capability threshold at 1B or
7B, so the effect of training tokens on reliance remains open; they nevertheless
fall on the same reliance--capability curve as the other families.

Taken together, on tasks that need step-by-step computation, scaling makes the
chain of thought \emph{more} load-bearing, not less. This supports chain-of-thought
monitoring in that regime and identifies direct-answer accuracy as the signal to
watch: if larger models begin to solve multi-step problems without writing the
steps, reliance on the chain may start to fall.

\bibliography{references}

\appendix
\section{Reproducibility Details}
\label{sec:appendix}
{\sloppy Table~\ref{tab:config} lists the full configuration and
Table~\ref{tab:edits} the per-model breakdown of edit outcomes. Per-model, per-$k$
and per-condition results, the generated items and the analysis scripts
(\texttt{step5\_score.py} and \texttt{step7\_paper.py}) are released with the
paper.\par}

\input{tables_appendix}

\end{document}

%% file: tables.tex
\begin{table*}[t]
\centering
\small
\setlength{\tabcolsep}{4.5pt}
\begin{tabular}{llrrrrrrr}
\toprule
Model & Stage & $N$ (B) & Prefix & CoT & Direct & Direct & Follows-edit & Follows-edit \\
 & & & acc. & $k{=}24$ & $k{=}2$ & $k{\ge}4$ & [95\% CI] & $\mid$ correct prefix \\
\midrule
\texttt{qwen3-0.6b-base} & base & 0.44 & 0.216 & 0.060 & 0.020 & 0.052 & 0.159 [0.142, 0.176] & 0.389 \\
\texttt{qwen3-1.7b-base} & base & 1.41 & 0.533 & 0.030 & 0.307 & 0.043 & 0.384 [0.361, 0.408] & 0.583 \\
\texttt{qwen3-4b-base} & base & 3.63 & 0.966 & 0.840 & 0.373 & 0.041 & 0.931 [0.919, 0.942] & 0.934 \\
\texttt{qwen3-8b-base} & base & 6.95 & 0.979 & 0.847 & 0.337 & 0.041 & 0.934 [0.923, 0.944] & 0.934 \\
\texttt{qwen3-14b-base} & base & 13.21 & 0.998 & 0.977 & 0.513 & 0.052 & 0.973 [0.966, 0.981] & 0.973 \\
\midrule
\texttt{qwen3-0.6b} & post & 0.44 & 0.225 & 0.027 & 0.087 & 0.043 & 0.178 [0.161, 0.196] & 0.435 \\
\texttt{qwen3-1.7b} & post & 1.41 & 0.934 & 0.447 & 0.233 & 0.050 & 0.854 [0.837, 0.870] & 0.861 \\
\texttt{qwen3-4b} & post & 3.63 & 0.962 & 0.850 & 0.463 & 0.048 & 0.913 [0.900, 0.926] & 0.912 \\
\texttt{qwen3-8b} & post & 6.95 & 0.888 & 0.693 & 0.447 & 0.051 & 0.941 [0.930, 0.952] & 0.941 \\
\texttt{qwen3-14b} & post & 13.21 & 0.999 & 0.997 & 0.583 & 0.039 & 0.996 [0.992, 0.998] & 0.996 \\
\midrule
\texttt{gemma3-1b} & post & 0.70 & 0.396 & 0.037 & 0.113 & 0.046 & 0.267 [0.246, 0.289] & 0.487 \\
\texttt{gemma3-4b} & post & 3.21 & 0.937 & 0.300 & 0.200 & 0.039 & 0.854 [0.838, 0.870] & 0.876 \\
\texttt{gemma3-12b} & post & 10.76 & 0.998 & 0.983 & 0.433 & 0.050 & 0.984 [0.978, 0.989] & 0.984 \\
\midrule
\texttt{llama3.2-1b} & post & 0.97 & 0.146 & 0.027 & 0.073 & 0.031 & 0.067 [0.056, 0.079] & 0.137 \\
\texttt{llama3.2-3b} & post & 2.82 & 0.544 & 0.010 & 0.000 & 0.002 & 0.288 [0.267, 0.308] & 0.427 \\
\texttt{llama3.1-8b} & post & 6.98 & 0.852 & 0.420 & 0.270 & 0.037 & 0.725 [0.704, 0.745] & 0.758 \\
\midrule
\texttt{qwen3-4b} (NF4) & post & 3.63 & 0.883 & 0.677 & 0.377 & 0.028 & 0.869 [0.854, 0.885] & 0.879 \\
\bottomrule
\end{tabular}
\caption{Per-model results on the box-swap task (1{,}800 problems, 7{,}200 items per model).
$N$: non-embedding parameters. \emph{Prefix acc.}: fraction of unedited CoT runs whose
states are correct up to the edit position. \emph{CoT}/\emph{Direct}: final-answer
accuracy with and without written steps; chance is $1/24\approx0.042$.
\emph{Follows-edit}: fraction of edited runs whose continuation and answer are
consistent with the corrupted state (95\% bootstrap CI over problems); the last
column restricts to problems on which the model's own unedited prefix was correct.
The NF4 row is a quantization ablation and is excluded from all fits.}
\label{tab:main}
\end{table*}
\begin{table}[t]
\centering
\small
\setlength{\tabcolsep}{4pt}
\begin{tabular}{llrrrr}
\toprule
Models & Metric & $n$ & $\rho_N$ & Midpoint & $R^2$ \\
\midrule
Qwen3 & Follows-edit & 10 & 0.985 & 1.07B & 0.892 \\
Qwen3 & Follows-edit $\mid$ prefix & 10 & 0.975 & 0.89B & 0.914 \\
Qwen3 & Prefix acc. & 10 & 0.886 & 0.84B & 0.902 \\
Qwen3 & CoT acc.\ $k{=}24$ & 10 & 0.886 & 2.19B & 0.884 \\
Qwen3 & Direct acc.\ $k{=}2$ & 10 & 0.911 & 9.61B & 0.843 \\
\midrule
3 families & Follows-edit & 16 & 0.896 & 1.67B & 0.737 \\
3 families & Prefix acc. & 16 & 0.866 & 1.28B & 0.787 \\
3 fam., capable & Follows-edit $\mid$ prefix & 12 & 0.797 & -- & -- \\
\bottomrule
\end{tabular}
\caption{Scaling of reliance and capability with model size. $\rho_N$: Spearman
correlation with $N$. Midpoint and $R^2$ are from a saturating logistic in
$\log_{10} N$, $y = \ell + (1-\ell)\,\sigma\!\left(a(\log_{10}N - b)\right)$; the
midpoint is $10^{b}$. ``3 families'' pools Qwen3, Gemma~3 and Llama~3; ``capable''
keeps models with prefix accuracy $\ge 0.5$ ($p=0.002$ for $\rho_N$).}
\label{tab:fits}
\end{table}
\begin{table}[t]
\centering
\small
\setlength{\tabcolsep}{4pt}
\begin{tabular}{rrrrrr}
\toprule
 & \multicolumn{2}{c}{Prefix acc.} & \multicolumn{3}{c}{Follows-edit $\mid$ correct prefix} \\
$N$ (B) & base & post & base & post & $\Delta$ \\
\midrule
0.44 & 0.216 & 0.225 & 0.389 & 0.435 & 0.046 \\
1.41 & 0.533 & 0.934 & 0.583 & 0.861 & 0.278 \\
3.63 & 0.966 & 0.962 & 0.934 & 0.912 & $-$0.022 \\
6.95 & 0.979 & 0.888 & 0.934 & 0.941 & 0.007 \\
13.21 & 0.998 & 0.999 & 0.973 & 0.996 & 0.023 \\
\bottomrule
\end{tabular}
\caption{Post-training effect within Qwen3. Reliance differs between base and
post-trained checkpoints only at 1.7B, where the two also differ sharply in
capability (prefix accuracy 0.533 vs.\ 0.934); from 4B upward the difference is
within $\pm0.03$.}
\label{tab:post}
\end{table}
\begin{table}[t]
\centering
\small
\setlength{\tabcolsep}{3.5pt}
\begin{tabular}{lrrrrr}
\toprule
Checkpoint & Tokens & Prefix & Direct & Follows & F.-e.\ $\mid$ \\
 & (T) & acc. & $k{=}2$ & edit & prefix ($n$) \\
\midrule
\texttt{olmo2-1b-336B} & 0.34 & 0.012 & 0.000 & 0.018 & 0.091 (22) \\
\texttt{olmo2-1b-944B} & 0.94 & 0.007 & 0.017 & 0.013 & 0.000 (12) \\
\texttt{olmo2-1b-1972B} & 1.97 & 0.006 & 0.017 & 0.017 & 0.091 (11) \\
\texttt{olmo2-1b-2978B} & 2.98 & 0.014 & 0.010 & 0.012 & 0.040 (25) \\
\texttt{olmo2-1b-main} & final & 0.009 & 0.027 & 0.007 & 0.176 (17) \\
\midrule
\texttt{olmo2-7b-1934B} & 1.93 & 0.028 & 0.030 & 0.031 & 0.000 (51) \\
\texttt{olmo2-7b-2924B} & 2.92 & 0.032 & 0.053 & 0.041 & 0.155 (58) \\
\texttt{olmo2-7b-main} & final & 0.198 & 0.217 & 0.128 & 0.303 (356) \\
\bottomrule
\end{tabular}
\caption{OLMo~2 intermediate pretraining checkpoints. ``final'' is the released
checkpoint ($\approx$4T pretraining tokens plus mid-training). No checkpoint
reaches the 0.5 capability threshold, so reliance cannot be estimated along the
token axis; the conditional column is shown for completeness but rests on very
few problems for all but \texttt{olmo2-7b-main}.}
\label{tab:olmo}
\end{table}

%% file: tables_appendix.tex
\begin{table}[t]
\centering
\small
\setlength{\tabcolsep}{5pt}
\begin{tabular}{ll}
\toprule
Setting & Value \\
\midrule
Task & 4-box swap tracking, letters A--D \\
Swaps per problem $k$ & 2, 4, 8, 12, 16, 24 \\
Problems & 300 per $k$ (1{,}800 total) \\
Items per model & 7{,}200 (4 conditions $\times$ 1{,}800) \\
Prompt & instruction + 3 worked examples ($k{=}2,3,4$) \\
Step format & \texttt{Step $t$ (swap $i$ and $j$): 1:C 2:B 3:A 4:D} \\
Chat models & native chat template \\
Qwen3 thinking mode & disabled (empty think block) \\
Decoding & greedy, stop at first \texttt{Answer:} line \\
Parser & line-based, reads up to the first answer line \\
Edit & state at step $m$ replaced by a different \\
 & valid arrangement; model continues from $m{+}1$ \\
Capability threshold & prefix accuracy $\ge 0.5$ (fixed in advance) \\
CIs & 1{,}000 bootstrap resamples over problems \\
Precision & 16-bit; 4-bit NF4 where a model \\
 & does not fit in 16\,GB \\
Hardware & one NVIDIA T4 (16\,GB), Azure ML \\
Training & none (evaluation only) \\
\bottomrule
\end{tabular}
\caption{Experimental configuration. All runs are inference-only and use a
checkpointed, preemption-tolerant job runner, so every reported number comes
from a complete set of 7{,}200 generations per model.}
\label{tab:config}
\end{table}

\begin{table}[t]
\centering
\small
\setlength{\tabcolsep}{4pt}
\begin{tabular}{lrrrr}
\toprule
Model & Follows & Repair & Wrong & Invalid \\
 & edit & & other & \\
\midrule
\texttt{qwen3-0.6b-base} & 0.16 & 0.04 & 0.77 & 0.03 \\
\texttt{qwen3-1.7b-base} & 0.38 & 0.05 & 0.56 & 0.01 \\
\texttt{qwen3-4b-base} & 0.93 & 0.01 & 0.06 & 0.00 \\
\texttt{qwen3-8b-base} & 0.93 & 0.01 & 0.05 & 0.00 \\
\texttt{qwen3-14b-base} & 0.97 & 0.00 & 0.02 & 0.00 \\
\midrule
\texttt{qwen3-0.6b} & 0.18 & 0.03 & 0.55 & 0.24 \\
\texttt{qwen3-1.7b} & 0.85 & 0.01 & 0.07 & 0.06 \\
\texttt{qwen3-4b} & 0.91 & 0.02 & 0.07 & 0.00 \\
\texttt{qwen3-4b-nf4} & 0.87 & 0.01 & 0.11 & 0.00 \\
\texttt{qwen3-8b} & 0.94 & 0.01 & 0.04 & 0.01 \\
\texttt{qwen3-14b} & 1.00 & 0.00 & 0.00 & 0.00 \\
\midrule
\texttt{gemma3-1b} & 0.27 & 0.01 & 0.59 & 0.11 \\
\texttt{gemma3-4b} & 0.85 & 0.00 & 0.12 & 0.02 \\
\midrule
\texttt{llama3.2-1b} & 0.07 & 0.03 & 0.84 & 0.06 \\
\texttt{llama3.2-3b} & 0.29 & 0.02 & 0.15 & 0.54 \\
\texttt{llama3.1-8b} & 0.72 & 0.02 & 0.12 & 0.13 \\
\midrule
\texttt{olmo2-1b-336B} & 0.02 & 0.01 & 0.49 & 0.46 \\
\texttt{olmo2-1b-944B} & 0.01 & 0.00 & 0.37 & 0.60 \\
\texttt{olmo2-1b-1972B} & 0.02 & 0.01 & 0.35 & 0.62 \\
\texttt{olmo2-1b-2978B} & 0.01 & 0.01 & 0.43 & 0.54 \\
\texttt{olmo2-1b-main} & 0.01 & 0.00 & 0.14 & 0.85 \\
\texttt{olmo2-7b-1934B} & 0.03 & 0.02 & 0.91 & 0.01 \\
\texttt{olmo2-7b-2924B} & 0.04 & 0.02 & 0.71 & 0.22 \\
\texttt{olmo2-7b-main} & 0.13 & 0.06 & 0.75 & 0.07 \\
\texttt{gemma3-12b} & 0.98 & -- & -- & -- \\
\bottomrule
\end{tabular}
\caption{Breakdown of outcomes on edited items. \emph{Repair}: the model returns to
the true trajectory; \emph{wrong other}: a parseable trajectory consistent with
neither the edit nor the truth; \emph{invalid}: no parseable answer. Rarer
categories (answer contradicting the model's own final step) are $\le0.02$ for every
model. For \texttt{gemma3-12b} only the follows-edit rate was logged in the final
scoring run; the full breakdown is in the released \texttt{table\_edit\_outcomes.csv}.}
\label{tab:edits}
\end{table}